\documentclass[letterpaper, conference, twocolumn, final]{ieeeconf}

\IEEEoverridecommandlockouts                              %

\usepackage{graphicx} 
\usepackage{booktabs}
\usepackage[dvipsnames]{xcolor}
\usepackage{amsfonts,amsmath,amssymb}
\usepackage[utf8]{inputenc}
\usepackage{epstopdf}
\usepackage{subfigure}
\usepackage{soul}
\usepackage{hyperref}
\graphicspath{{figs/}}
\usepackage{todonotes}
\usepackage{stfloats}
\usepackage{enumerate}

\newcommand{\nbrs}{\mathcal{N}} 
 \newcommand{\real}{\mathbb{R}}

\newcommand{\until}[1]{\{1, \ldots, #1\}}

\newcommand{\pos}{\mathrm{p}} 
\newcommand{\vel}{\mathrm{v}}

\usepackage{listings}
\definecolor{mygray}{gray}{0.95}
\makeatletter
\newcommand\footnoteref[1]{\protected@xdef\@thefnmark{\ref{#1}}\@footnotemark}
\makeatother

\usepackage{cuted}

 \def \packagename/{\textsc{CrazyChoir}} %
\title{CrazyChoir: Flying Swarms of Crazyflie Quadrotors in ROS 2}
\author{%
Lorenzo Pichierri, Andrea Testa, Giuseppe Notarstefano \thanks{ This work was
supported in part by the Italian Ministry of Foreign Affairs and International
Cooperation, grant number BR22GR01.
  }%
  \thanks{Authors are with the Department of Electrical, Electronic and
  Information Engineering, University of Bologna, Bologna, Italy.
  \texttt{\{lorenzo.pichierri, a.testa, giuseppe.notarstefano\}@unibo.it}.}%
}
\begin{document}

\maketitle

\begin{strip}\leavevmode\kern15pt
  \begin{minipage}{\dimexpr\linewidth-30pt\relax}
  {\vspace{-2.5cm} \bf \textcopyright 2023 IEEE. Personal use of this material
  is permitted.  Permission from IEEE must be obtained for all other uses, in
  any current or future media, including reprinting/republishing this material
  for advertising or promotional purposes, creating new collective works, for
  resale or redistribution to servers or lists, or reuse of any copyrighted
  component of this work in other works.}
  \end{minipage}
  \end{strip}

\begin{abstract} %
  This paper introduces \packagename/, a modular Python framework 
  based on the Robot Operating System (ROS) 2. The toolbox provides a
  comprehensive set of functionalities to simulate and run experiments on teams
  of cooperating Crazyflie nano-quadrotors.
Specifically, it allows users to perform realistic simulations over robotic
    simulators as, e.g., Webots and includes bindings of the firmware control
    and planning functions. The toolbox also provides libraries to perform radio
    communication with Crazyflie directly inside ROS~2 scripts.
  The package can be thus used to design, implement and test planning strategies
  and control schemes for a Crazyflie nano-quadrotor. Moreover, the modular
  structure of \packagename/ allows users to easily implement online distributed
  optimization and control schemes over multiple quadrotors.
The \packagename/ package is validated via simulations and experiments on a
    swarm of Crazyflies for formation control, pickup-and-delivery vehicle
    routing and trajectory tracking tasks.
\packagename/ is available at \url{https://github.com/OPT4SMART/crazychoir}.
\end{abstract}


\section{Introduction}
\label{sec:introduction}
\begin{figure}[ht]
	\centering
	\includegraphics[width=.85\columnwidth]{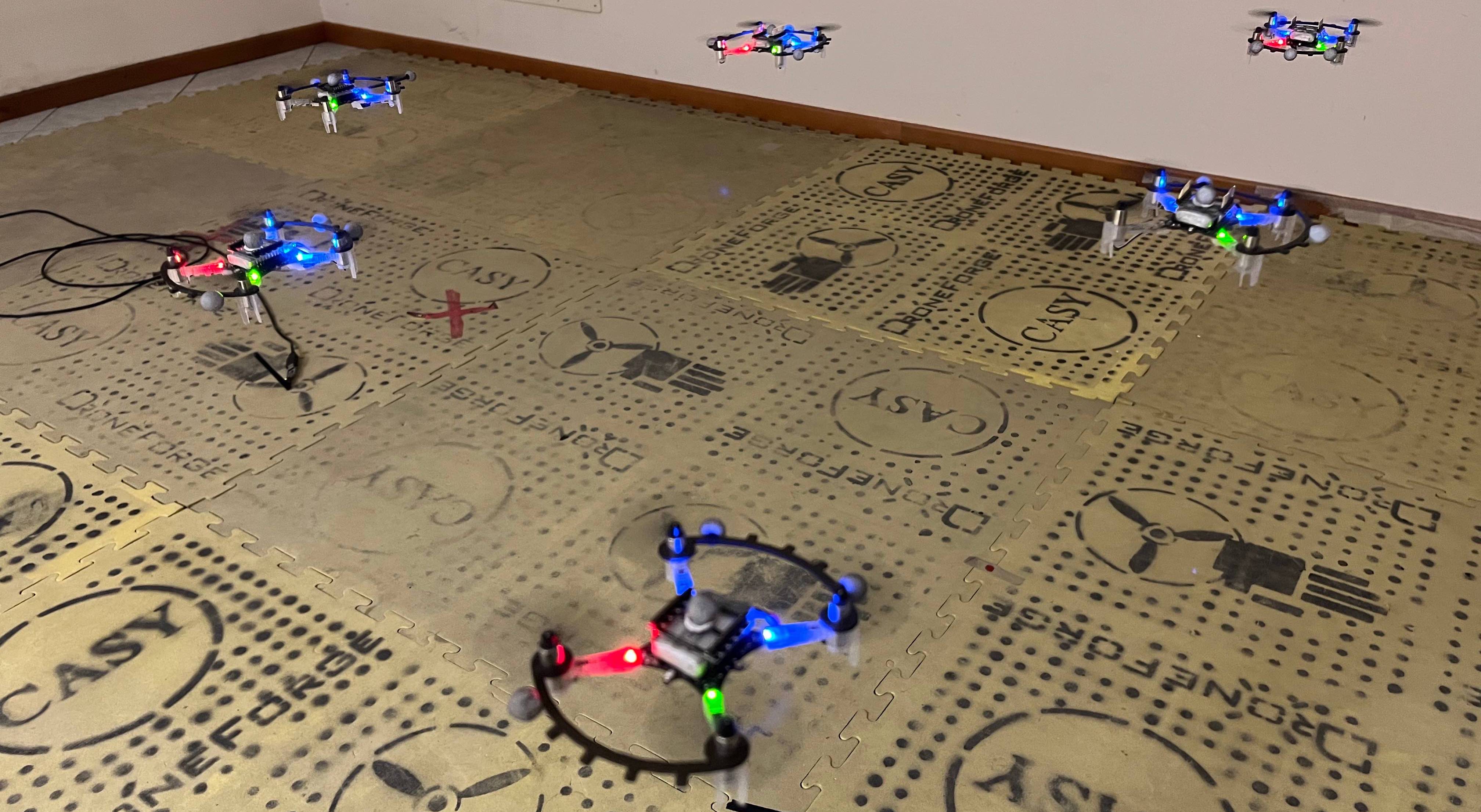}
	\caption{Snapshot of a swarm of Crazyflie nano-quadrotors in our testbed.}
	\label{fig:Crazyflie}
\end{figure}

UAV swarms have gained a great interest in both research and industrial
  applications due to their applicability in a wide range of scenarios.  
In the last years the Crazyflie platform (Figure~\ref{fig:Crazyflie}) has gained
a lot of attention among researchers. Crazyflie is a cheap nano-quadrotor
weighting $27 {\rm\; g}$. It is endowed with an ARM Cortex-M4 microcontroller
and can receive radio messages through a so-called Crazyradio dongle. In the
last years, several efforts have been put into the development of simulation and
control toolboxes based on the Robot Operating System
(ROS)~\cite{quigley2009ros}. Recently, ROS is being substituted by ROS~2, which
is expanding its capabilities with novel
functionalities~\cite{maruyama2016exploring}. In this paper, we introduce
\packagename/, a ROS~2 package allowing researchers to run realistic simulations
and laboratory experiments on a swarm of cooperating Crazyflie nano-quadrotors.

\subsection{Related Work}
Control of quadrotors is currently performed with toolboxes based on the first
ROS version. As for toolboxes specific for the Crazyflie, the toolboxes
in~\cite{preiss2017crazyswarm,honig2017flying} allow user to run experiments on
multiple Crazyflies.  Researchers also developed tools to efficiently simulate
generic, autonomous quadrotors in a realistic environment.  An Unreal Engine
simulator for unmanned vehicles is proposed in~\cite{shah2018airsim}. Authors
in~\cite{song2020flightmare} propose a simulator based on Unreal Engine with
bridges for RotorS and OpenAI-Gym as to perform Reinforcement Learning tasks.
Works in~\cite{grabe2013telekyb,meyer2012comprehensive} propose two toolboxes to
simulate and control multiple UAVs. Finally, a ROS-Gazebo simulator for the
Erle-Copter is provided in~\cite{kumar2020erle}.  As for simulation toolboxes
tailored for the Crazyflie, authors in~\cite{silano2018crazys} extend the
well-known RotorS toolbox~\cite{furrer2016rotors} to target the Crazyflie. A
Gazebo-ROS-Simulink toolbox for the Crazyflie is proposed instead
in~\cite{nithya2019gazebo}.  However, in the above toolboxes the swarm is
handled by few, centralized ROS processes that interact with all the quadrotors,
thus affecting the scalability and robustness of the swarm. Moreover, they do
not involve distributed or decentralized design tools for swarm management and
control.  Recently, the novel ROS~2~\cite{macenski2022robot} is substituting
ROS. Indeed, ROS~2 allows multi-process communication leveraging the popular
Data Distribution Service (DDS) open standard. The DDS does not require a
centralized broker to dispatch messages, and nodes can implement self-discovery
procedures. Moreover, the DDS allows for the implemention of different Quality
of Service profiles, and provides reliable and secure communication. Also, ROS~2
has been designed for industrial settings, thus supporting
real-time~\cite{puck2020distributed} and embedded
systems~\cite{belsare2023micro} development.
Authors in~\cite{erHos2019ros2} propose a framework for collaborative industrial
manipulators. The work in~\cite{reke2020self} proposes a ROS~2-based
architecture for self-driving cars. As for works concerning software for
multi-robot systems in ROS~2, papers~\cite{kaiserros2swarm,mai2022driving}
address the development of toolboxes for swarm robotics on ground mobile robots.
Moreover, these works neglect cooperation among robots or simulate it via
(computationally demanding) all-to-all communication. None of these frameworks
is however tailored for the Crazyflie. Recently, Crazyswarm2~\cite{crazyswarm2}
has been proposed as a ROS~2 extension to
Crazyswarm~\cite{preiss2017crazyswarm}. Still, this package does not include
routines to implement distributed optimization and control schemes on
cooperating Crazyflies. 
Finally, the toolbox in~\cite{testa2021choirbot} provides distributed
optimization and communication functionalities for teams of generic robots. This
package however does not handle Crazyflie swarms and radio communication with
the quadrotors. Moreover, as we detail next, our package provides simulations by
combining the Webots engine~\cite{michel2004cyberbotics} and Crazyflie firmware
bindings.

\subsection{Contributions}

\begin{figure*}[t]
  \centering
    \vspace*{5mm}
    \includegraphics[width=.75\textwidth]{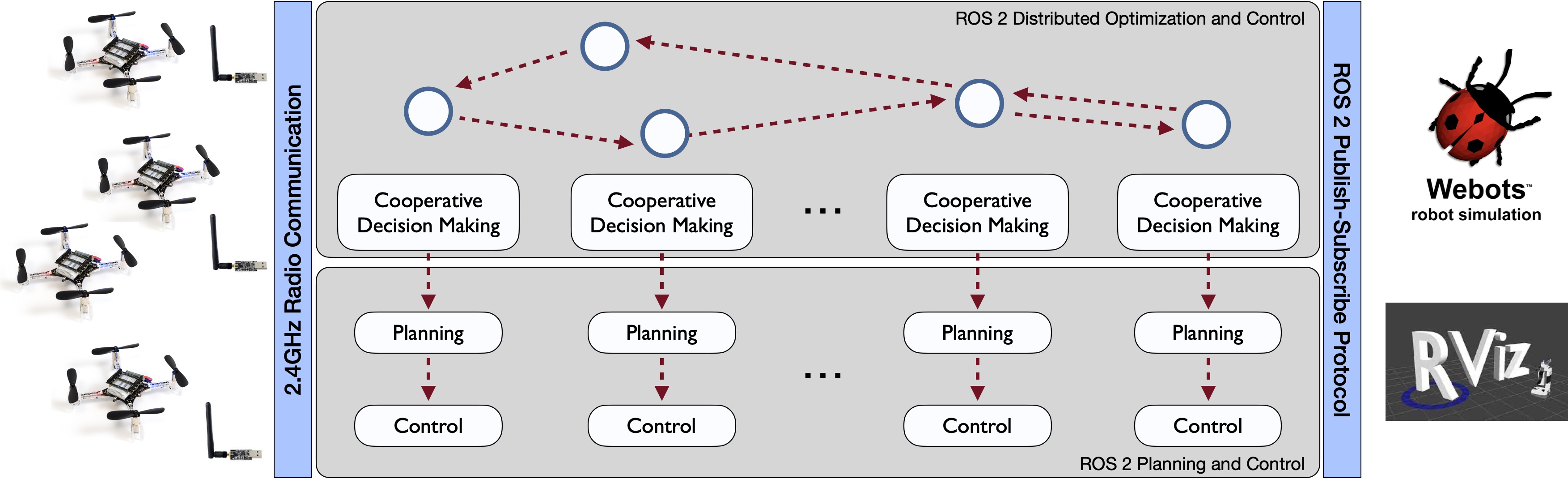}
    \caption{\packagename/ architecture. Crazyflies can exchange information
    with neighbors to implement distributed feedback and optimization schemes.
    Local classes handle planning and low-level control. Control inputs can be
    sent to simulation environments (Webots or \textsc{Rviz}) or to real robots
    via radio.}\label{fig:architecture}
  \end{figure*}

In this paper we introduce \packagename/, a ROS~2 modular framework to perform
realistic simulations and experiments on swarms of cooperating Crazyflie
nano-quadrotors.
The toolbox provides a comprehensive set of Python modules allowing for fast
prototyping of
(distributed) decision-making and control schemes on Crazyflie swarms.
  Specifically, \packagename/ is a scalable package in which each Crazyflie is
  handled by a set of independent ROS~2 processes, thus reducing single points
  of failure. These nodes are executed on a ROS-enabled workstation and can be
  allocated among multiple workstations connected on the same networks. Also,
  its modular structure allows the user to easily switch from simulative
  scenarios to real experiments by changing a few lines of code.
The toolbox is integrated with the Webots engine, so that users can perform
realistic simulations of multiple Crazyflie. To the best of the authors'
knowledge, realistic, CAD-based simulation of large-scale swarms is often
neglected.
Moreover, the simulator incorporates the firmware bindings provided by Bitcraze
implementing Crazyflie onboard control functions.  
Thereby, the user can perform parameter tuning in a safe simulation environment.
The proposed framework also allows for lightweight simulations by means of an
ad-hoc numerical integrator combined with the \textsc{Rviz} visualizer. Finally,
\packagename/ supports \textsc{DISROPT}~\cite{farina2020disropt} and
\textsc{ChoiRbot}~\cite{testa2021choirbot} toolboxes, thus allowing users to
implement distributed optimization schemes and handle inter-robot communications
over ROS~2 according to a user-defined graph.
The paper unfolds as follows. In Section~\ref{sec:architecture}, we detail the
architecture of \packagename/. Section~\ref{sec:cf-control} details libraries
for control schemes. Section~\ref{sec:distributed} discusses swarm 
decision making strategies, distributed control and distributed optimization
functionalities of the proposed package. In Section~\ref{sec:radio}, we discuss
the communication bridge with the Crazyflie. Section~\ref{sec:integration} shows
how to simulate swarms of Crazyflies in Webots and \textsc{Rviz}. In
Section~\ref{sec:simulation}, we provide a use-case implementation, while other
experiments are discussed in Section~\ref{sec:experiment}.

\section{Architecture Description}
\label{sec:architecture}
\packagename/ is written in Python and is based on the novel ROS~2 framework. In
\packagename/, each Crazyflie is handled by a set of ROS~2 processes (also
called \emph{nodes}). These nodes handle simulation, control, planning and radio
communication for each quadrotor, as well as inter-robot communication and
cooperation. In the proposed package, these functionalities are implemented in a
modular fashion as a set of Python classes. At runtime, these classes are
instantiated in a dedicated ROS~2 node as a standalone process. 
An illustrative representation of the software architecture is in
Figure~\ref{fig:architecture}.
The software architecture of \packagename/ is made of five main blocks:
	$i)$ control layer, $ii)$ swarm planning layer, $iii)$ cooperative decision
	making layer, $iv)$ radio communication layer, $v)$ realistic simulation
	layer.
We now briefly introduce the main components of the proposed architecture. A
detailed description is provided in the following sections.
The control layer provides a set of classes that can be extended to implement
off-board feedback-control schemes. 
These control classes receive reference trajectories from the planning module,
in which users can also provide graphical inputs to describe complex
trajectories.
The trajectory planning functionalities are used to generate feasible
trajectories, steering the generic quadrotor to fulfill high-level decisions.
These decisions are taken in another module provided by \packagename/, in which
users can implement cooperative, decision-making and distributed feedback
schemes on swarms of Crazyflies. This cooperative decision making layer indeed
provides functionalities to implement distributed, online optimization and
control schemes.
Leveraging this cooperative 
framework, robots can jointly fulfill complex tasks by exchanging messages with
neighboring robots.
The control inputs generated by the control classes, or the setpoints provided
by the planning module, 
can either be sent to the Crazyflie via radio, or transmitted via ROS~2 topics
to the simulators provided in \packagename/.
In particular, users can choose between realistic simulations, developed in
Webots, and lightweight numerical integrations, visualized in \textsc{Rviz}.
Finally, \packagename/ can be interfaced with motion capture systems as, e.g.,
Vicon so that users can retrieve quadrotor poses during the experiments. The
proposed package implements derivative and low-pass filters to obtain linear and
angular velocities for each robot. Indeed, motion capture systems usually
provide only position and attitude data.
Thanks to this modular design, users can combine classes from the different
layers according to their needs. 
As we detail in the following, this allows for switching from simulations to
experiments with few lines of code.
Users can quickly extend the proposed modules with novel functionalities and,
thanks to the ROS~2 capabilities, can also deploy the software on different
workstations.

\section{\packagename/ Control Library}
\label{sec:cf-control}
\begin{figure}[htpb]
\centering
\includegraphics[width=.7\columnwidth]{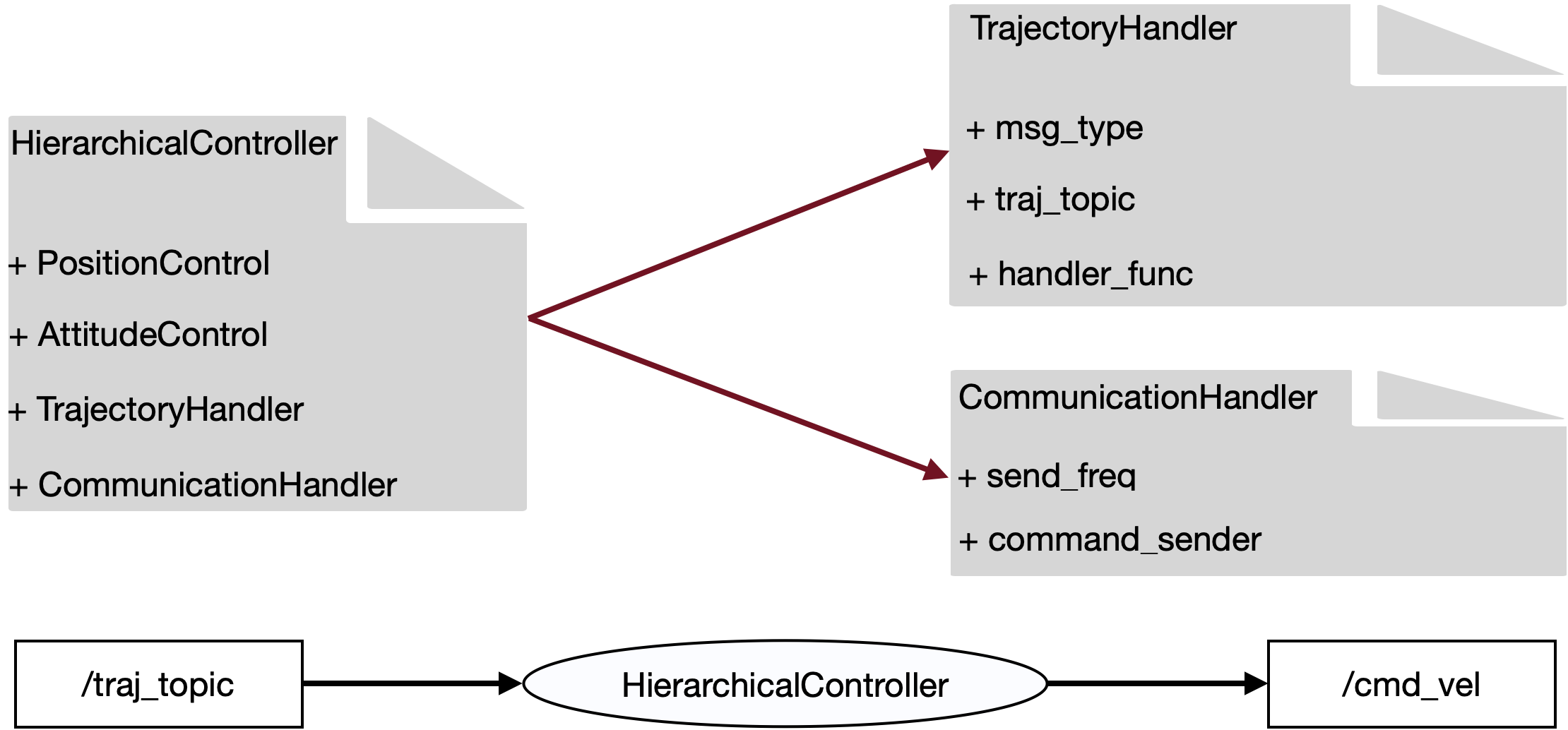}
\caption{\texttt{HierarchicalControl} class and involved topics.}
\label{fig:controller}
\end{figure}

\packagename/ provides a \texttt{CrazyflieController} template class that can be
specialized to implement the desired control schemes for the Crazyflie. This
class aims at mapping desired trajectory inputs to a control setpoint to be sent
to the Crazyflie or to the simulators. The \texttt{CrazyflieController} class is
designed according to the \emph{Strategy} pattern~\cite{gamma1995design} so that
the specific control law can be defined at runtime. %
Specifically, the class owns as attributes a set of classes specifying its
behavior. That is, it owns \texttt{TrajectoryHandler} and
\texttt{CommunicationHandler} classes specifying how desired trajectories will
be published to the controller and which control input must be communicated to
the Crazyflie. In the proposed package, we already provide a
\texttt{HierarchicalController} class that implements classical flatness-based
control schemes~\cite{mellinger2011minimum}. This class is suitable to track a
desired flat-output trajectory (e.g., position and yaw profiles with their
derivatives), or it can be used to track desired acceleration profiles coming
from, e.g., double-integrator distributed control laws. We provide an example of
this setting in Section~\ref{sec:simulation}, in which we use this class to
perform a distributed formation control task. \texttt{HierarchicalController}
extends the \texttt{CrazyflieController} template. To this end, it owns two
additional classes \texttt{PositionControl} and \texttt{AttitudeControl},
implementing position and angular control, respectively. 
The outputs of the \texttt{PositionControl} class are the thrust and desired
attitude profile. 
The \texttt{AttitudeControl} classes instead generate an angular velocity
profile given desired attitude references. As an example, we implement the
geometric control law in~\cite{lee2010geometric}. 
We refer the reader to the next sections for a detailed discussion of these
modules. A sketch of the role of these classes and the related topics is
provided in Figure~\ref{fig:controller}. Here, the arrows 
generated from the \texttt{HierarchicalController} class represent the fact that
\texttt{TrajectoryHandler} and \texttt{CommunicationHandler} are attributes of
\texttt{HierarchicalController}, that call their methods. The
\texttt{HierarchicalController} class receives trajectory messages on
\texttt{/traj\_topic} and sends control messages on \texttt{/cmd\_vel} topic.

\section{\packagename/ Modules for Cooperative Decision Making and Planning}
\label{sec:distributed}
\packagename/ provides modules to perform off-board planning tasks for each
quadrotor. Also, the proposed package allows users to implement cooperative
decision-making and distributed feedback schemes on swarms of Crazyflies.

\subsection{GUI-based Swarm Trajectory Planning}
\label{subsec:gui}

\begin{figure}[h]
  \centering
\includegraphics[width=.9\columnwidth]{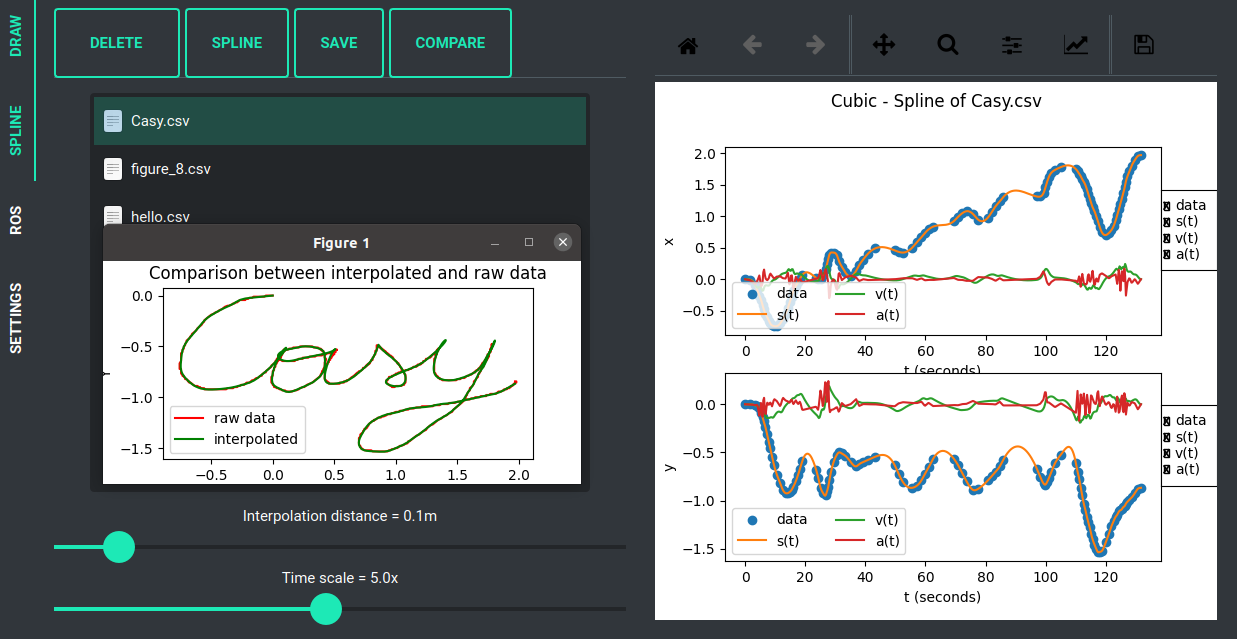}
	\caption{Snapshot of the proposed GUI.}
	\label{fig:gui}
\end{figure}

\packagename/ involves a set of functionalities to handle planning tasks. More
in detail, \packagename/ includes a graphical interface to provide trajectories
to a set of quadrotors.  In fact, one of these functionalities allows users to
hand-draw trajectories and directly send them to the chosen quadrotor(s). The
points of these hand-drawn trajectories are interpolated, and a polynomial
spline is constructed in order to ensure the continuity of velocity and
acceleration profiles. Then, users can tune the trajectory time in order to
increase/decrease its execution and avoid high accelerations. Also, this process
can be extended in order to use different planning methods and, e.g., manually
impose bounds on the trajectories. A snapshot of the interface is reported in
Figure~\ref{fig:gui}, where we draw the name of our laboratory (i.e.,
\emph{Casy}).  Further, in Section~\ref{sec:experiment} we present an example in
which a Crazyflie follows this trajectory.

This functionality can be exploited in scenarios in which, e.g., there is a
subset of \emph{leading} quadrotors that move in the space and a subset of
\emph{following} quadrotors that run cooperative algorithms to suitably track
the leaders.
In Section~\ref{sec:simulation}, we provide an example of formation control with
moving leaders to highlight this aspect.
The GUI can be also used to help the final user to manage classical operations
needed during simulations/experiments (e.g., hovering, landing, starting the
experiment, etc.).
Then, the package provides methods to perform point-to-point trajectories and to
evaluate polynomial paths passing through multiple, arbitrary points.
These modules allow the user to send (e.g., to the controller node) position,
velocity, and acceleration setpoints constituting a smooth trajectory at the
desired communication rate. These classes also handle dynamic replanning
scenarios where a new trajectory has to be evaluated while the quadrotor is
already following another path.
To interface planning classes with, e.g. controller classes, \packagename/
provides a \texttt{TrajectoryHandler} Strategy class. This class details the
topic on which the desired trajectory will be published. 
In this way, ROS~2 nodes dynamically instantiate the proper ROS~2 publisher and
subscribers. The class also specifies the trajectory message type and a callback
function to extract the trajectory from the message. 

\subsection{Distributed Optimization and Control}
\label{subsec:distr-opt}
In \packagename/, the user can implement \emph{distributed} optimization and
control schemes in which teams of quadrotors exchange suitable data in order to
solve complex tasks. In this cooperative setting, each quadrotor communicates
with few, neighboring quadrotors, possibly according to time-varying
communication topologies. As a result, the team of robots leverages the
\emph{local} knowledge of each quadrotor (coming, e.g. from onboard sensors) and
exploits inter-robot communications to achieve a global goal.
\packagename/ implements a set of functionalities to deploy distributed
optimization and control schemes in which users can implement distributed
algorithms from the perspective of each robot. Embedding
DISROPT~\cite{farina2020disropt} functionalities, users can model different
optimization problems in which each robot knowns only a part of the objective
function and constraints, eventually including non-convex, mixed-integer sets.
An example is provided in Section~\ref{sec:pdvrp}. Inter-robot communication is
then handled using the ROS~2 middleware. To this end, it is compatible with
\textsc{ChoiRbot}~\cite{testa2021choirbot}, a ROS~2 package for cooperative
robotics.
More in detail, it allows users to implement static or time-varying directed
communication networks. This is performed according to the publisher-subscriber
protocol.
Exploiting the Quality of Service feature introduced in ROS~2, the package
allows for the implementation of synchronous and asynchronous schemes, as well
as lossy communications.
This set of functionalities is provided by two main classes, \texttt{CFGuidance}
and \texttt{CFDistributedFeedback}, that extend \textsc{ChoiRbot}
functionalities in order to implement distributed schemes tailored for Crazyflie
swarms. 
The two classes expose to users a set of pre-defined methods \texttt{send},
\texttt{receive}, \texttt{asynchronous\_receive}, \texttt{neighbors\_exchange}
(to simultaneously perform send/receive) according to some user-defined
communication graph. These functions also allow to send different kinds of data
to different neighbors.  An interesting feature is that the user does not need
to define any ROS message.  Indeed, all the data is automatically serialized and
sent as a \texttt{std\_msgs/ByteMultiArray} on the sender side, while it is
de-serialized on the receiver side.  Thus, users can extend these classes
according to their needs by only implementing the desired algorithmic strategies
without warning about inter-robot communications.  We remind the reader to
Section~\ref{sec:simulation} and Section~\ref{sec:experiment} for usage examples
of these classes for cooperative control and optimization tasks.

\section{Radio Communication}
\label{sec:radio}
\begin{figure}[h!]
\centering
\includegraphics[width=.8\columnwidth]{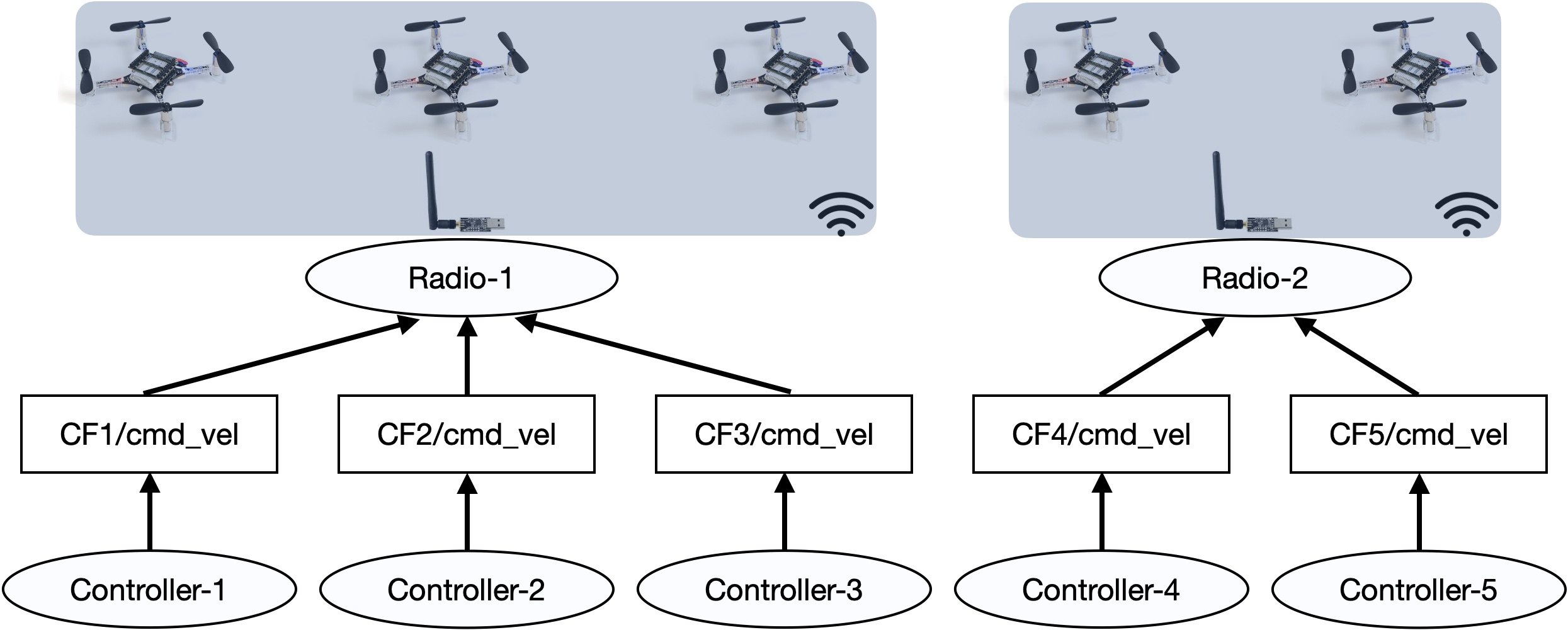}
\caption{Examples of nodes and topics involved in the radio communication for a swarm of $5$ quadrotors and $2$ radio dongles.}
\label{fig:radio}
\end{figure}

To physically control the Crazyflie nano-quadrotor, angular or angular-rate set
points can be sent via radio. Position or velocity set-points can be
communicated as well, together with periodical ground-truth information.
Crazyflies are equipped with an nRF51822 radio controller, used to communicate
with a PC by means of the Crazyradio PA. The Crazyradio PA is a $2.4 \;{\rm
GHz}$ USB radio dongle based on the nRF24LU1+. It transmits $32$-byte packets
with datarate up to $2$Mbits, spreading  
in a range of $125$ channels. 
Each Crazyflie is able to receive data on one of these channels at a certain
datarate. The choice of these communication parameter is crucial when performing
experiments with a large number of quadrotors. Indeed, a wrong choice may result
into a noisy communication channel and poor flight performance. To mitigate this
aspect, \packagename/ provides a script that identifies the channels that are
less subject to external interferences. To handle the communication with the
quadrotors, \packagename/ implements a tailored \texttt{RadioHandler} class.
Each Crazyradio PA is handled by an instantiation of this class into an
independent ROS~2 node. In particular, this class allows each dongle to
communicate with multiple Crazyflies by exploiting the Python methods of
\texttt{Crazyflie-lib-python} developed by Bitcraze.
The \texttt{RadioHandler} requires a list of \texttt{URI\_address}, used to set
a unique communication link with each Crazyflie associated with the radio. 
In an initialization procedure, the class updates a set of required parameters
of the Crazyflie related to the onboard controller. After that, each
\texttt{radio} node subscribes to the multiple \texttt{cmd\_vel} topics
published by the control nodes (cf. Section~\ref{sec:cf-control}) for each
Crazyflie associated with the dongle.
The subscription callback triggers the \texttt{cmd\_sender}, which is
implemented in subclasses of \texttt{RadioHandler}, and it is in charge of
elaborating the control input and forward it to the nano-quadrotors.
A sketch of this architecture is provided in Figure~\ref{fig:radio}.

We provide a set of subclasses that extend \texttt{RadioHandler}, namely
\texttt{RadioHandlerFPQR}, \texttt{RadioHandlerFRPY} and
\texttt{RadioHandlerXYZ}.
More in detail, the first one sends to the nano-quadrotors thrust and
angular-rate setpoints. The onboard firmware then evaluates a suitable control
input using a PID controller, comparing the setpoint with gyroscope data. 
The second one instead sends thrust and angular setpoints. An onboard PID
controller tracks these setpoints. The class leverages Bitcraze logging methods
to retrieve onboard log data.
These subclasses are tailored for precise control schemes and require control
inputs to be sent at high communication rates. The \texttt{RadioHandlerXYZ}
class instead sends position setpoints and ground truth information to the
robots. These setpoints are handled onboard by the
\texttt{higher\_level\_commander} implemented on the Crazyflie firmware. The
communication of position setpoints can be performed at a low communication
rate, in the order of a few Hertz. This fact reduces the number of messages on
the communication links. Thus, this class is tailored for experimental scenarios
with several quadrotors.
Each \texttt{radio} node handles a group of Crazyflies and thus requires a list
of radio URIs and a set of numerical identifiers to subscribe to the correct
topics.

\section{Simulation Tools in \packagename/}
\label{sec:integration}
In this section, we describe how to simulate a swarm of Crazyflies in
\packagename/. First, we detail how to perform this task in conjunction with
Webots, a realistic engine tailored for robot simulations. Then, we detail how
to run a numerical simulation using a custom integrator and  \textsc{Rviz}
visualization.
Although we focus on Webots, users can extend the package to leverage other
engines such as Gazebo. 

\subsection{Realistic Simulations via Webots }
\label{subsec:webots}
\packagename/ includes a set of functionalities, in the form of \emph{plugins},
to interact with Webots. These plugins receive inputs from the higher-level
classes (e.g., control inputs or waypoints coordinates) and map them to motor
commands. The command mapping is evaluated using Python bindings of the
Crazyflie firmware functions provided by Bitcraze. This allows the designer to
have realistic feedbacks on the quadrotor behavior. Also, the designer could use
this feature to test new firmware functions, e.g., embed control algorithm
directly onboard.
To properly simulate a robot in the Webots environment, also called
\textit{World}, the designer has to specify (i) the geometry of the robot, (ii)
external features that enrich the model (e.g. sensors, cameras, and actuators)
and (iii) plugins to control the actuators. 
The first requirement is achieved by importing a \texttt{.stl} file
(\textit{STereo Lithography} interface format). External features can be endowed
by filling up a \texttt{.urdf} file (\textit{Unified Robot Description Format}). 
Finally, to satisfy the third requirement, \packagename/ provides a class of
plugins to control the Crazyflie actuators.
Hence, to endow our package with an exhaustive set of Webots plugins, we
developed a parent class named \texttt{MotorCtrl}. At the beginning, the class
sets out the whole Crazyflie physical equipments, making available GPS
measurements and IMU detections. Moreover, the class initializes ROS~2
publishers and subscribers in order to interface Webots with  \packagename/
planners and controllers.
The main class of plugins \texttt{MotorCtrl} is extended by two subclasses. The
first subclass, \texttt{MotorCtrlFPQR}, takes as input references the desired
thrust and angular rates. The computation of these quantities is explained in
Section~\ref{sec:cf-control}. The mapping to motor torques is then performed
using firmware Python bindings, thus using the same model implemented on the
Crazyflie.
The second subclass, \texttt{MotorCtrlXYZ}, leverages the functionalities of the
trajectory planner coded in the firmware. Indeed, it receives desired position
and yaw setpoints that are forwarded to the planner methods to evaluate a
trajectory. As a result, the trajectory is elaborated by firmware bindings and
motor commands are provided to the simulated Crazyflie.

\subsection{Lightweight Simulations via Numerical Integration}
\label{sec:simulator}
To allow users to perform simulations without additional, external software, we
provide a dynamics simulator for the Crazyflie nano-quadrotor running an
Explicit Runge-Kutta method. The integration is performed at $100 \;{\rm Hz}$,
and is based on the following dynamics
\begin{align} 
\label{eq:first_dyn}
\dot{\pos} &= \vel \\
\label{eq:acceleration}
\dot{\vel} &= g z_W - \frac{u_1}{m}R(\eta)z_W - \frac{1}{m} R(\eta) A R(\eta)^\top \vel \\
\label{eq:last_dyn}
\dot{\eta} &= [u_2 \; u_3 \; u_4]^\top
\end{align}
Here, $\pos \in \real^3$ is the position of the quadrotor in the world frame
$\mathcal{F}_W=\{x_W, y_W, z_W\}$, $\vel \in \real^3$ is its linear velocity,
and $m=0.027 \;{\rm Kg}$ is the Crazyflie mass. 
The rotation matrix is denoted by $R \in SO(3)$, and the vector $\eta= [\varphi,
\theta, \psi]^\top$ stacks the Euler's angles, namely, roll, pitch and yaw. The
Crazyflie allows the user to send thrust, roll rate, pitch rate and yaw rate as
control actions. Thus, the control inputs in the considered dynamics are
collected by the vector $u = [u_1 \; u_2 \; u_3 \; u_4]^\top\in\real^4$, with
$u_1 \in \real$ denoting the thrust and $u_2,u_3,u_4 \in \real$ the angular
rates (according to the $Z-Y-X$ extrinsic Euler representation).
The term $RAR^\top \vel$ in~\eqref{eq:acceleration} refers to the rotor drag
effect as modelled in~\cite{kai2017nonlinear}. The numerical values of the drag
parameters have been chosen according to the discussion
in~\cite{forster2015system}.
Thanks to the modular structure of the package, the user can extend or modify
the proposed integrator to simulate more complex dynamics.
To visualize the quadrotor motion, we also provide a visualization utility based
on \textsc{Rviz}.
\section{\packagename/ Hands-on Example: Bearing-Based Formation Control}
\label{sec:simulation}
This section aims at introducing the user to our toolbox. To this end, we show
how to simulate the bearing-based distributed formation control scheme as
in~\cite{zhao2017translational} over a swarm of Crazyflie. We start by providing
the problem set-up and the classes needed for its implementation. Then, we show
how to run the simulation both in \textsc{Rviz} and Webots. Finally, we show how
\packagename/ allows users to easily switch from simulation to experiment and
run the same setting on a real swarm of Crazyflies.

\subsection{Problem Set-up}
We consider a network of $N$ quadrotors, divided into $N_l$ leaders and $N-N_l$
followers. The objective is to control the followers to deploy a desired
formation in the space, while the leaders stand still in their positions. In the
considered distributed set-up, each quadrotor can communicate with a set
$\nbrs_i$ of neighboring quadrotors.
The formation is defined by a set of \emph{bearings} $g^\star_{ij}$
for all the couples $(i,j)$ with $i\in\until{N}$ and $j\in\nbrs_i$. 
Authors in~\cite{zhao2017translational} consider double-integrator systems in
the form $\ddot{\pos}_i^{di} = u_i^{di}$. 
Leaders apply $u_i^{di} = 0$, while followers implement %
\begin{align}
\label{eq:bearing_fc}
	u_i^{di} = -\textstyle\sum_{j \in \nbrs_i} P_{g_{ij}^{\star}}[k_{\pos}(\pos_i - \pos_j) + k_{\vel}(\vel_i - \vel_j)],
\end{align}
where $P_{g_{ij}^{\star}} = I_3 - g^\star_{ij}g^{\star \top}_{ij}$ and $I_3$ is
the $3\times3$ identity matrix. Notice that this input cannot be directly fed to
the quadrotors. In our example, we use the control input $u_i^{di}$ as a desired
acceleration profile for the $i$-th Crazyflie. This profile is then tracked
using a flatness-based controller (cf. Section~\ref{sec:cf-control}).

\subsection{Software Implementation}
\label{subsec:simulation-software}
\begin{figure}[h!]
\centering
	\includegraphics[width=.9\columnwidth]{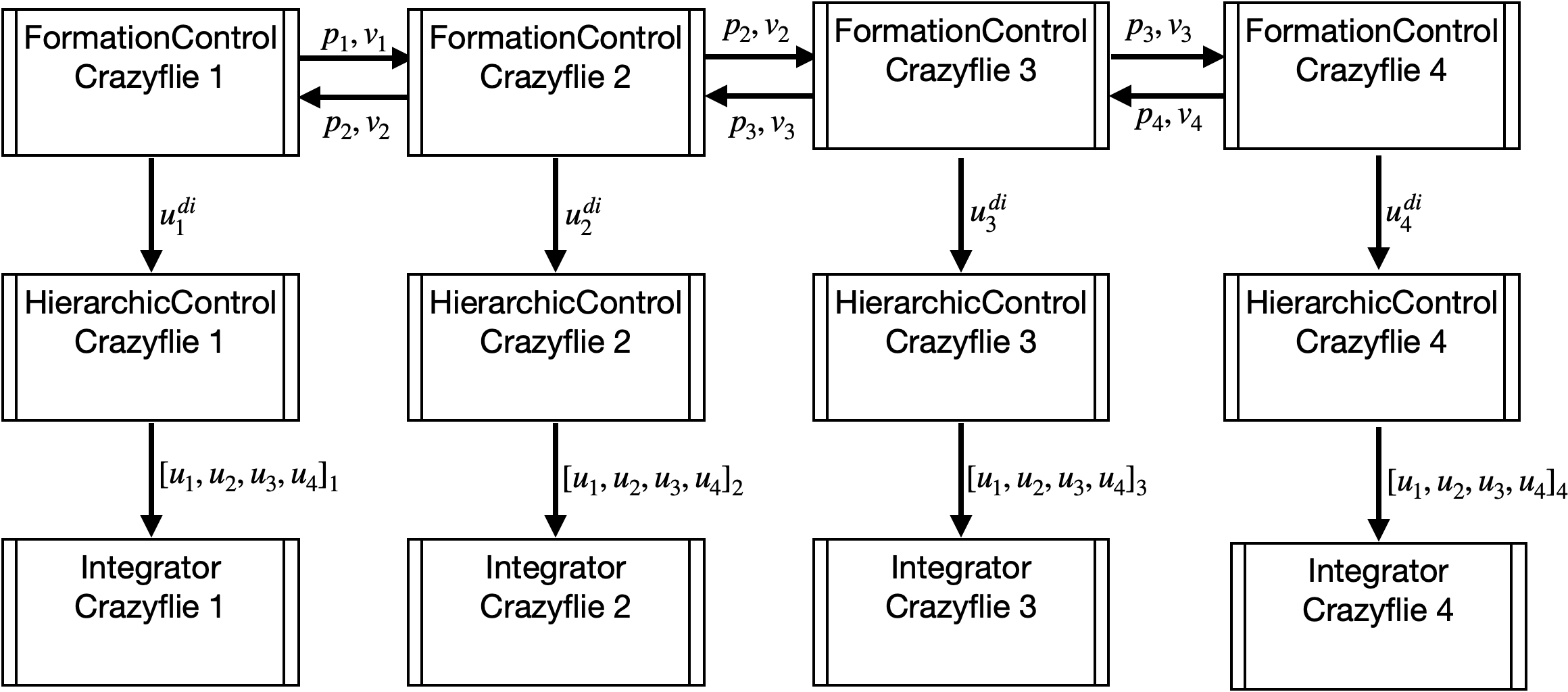}
	\caption{Class communication diagram.  Rectangles represent different classes,
 each one handled by a different node. Arrows represent directional
 communications, which are performed leveraging ROS~2 topics. On each arrow,
 exchanged data is reported.}
	\label{fig:class_comm}
\end{figure}

To implement the distributed bearing-based formation control scheme, we need
three classes. The first one implements the distributed control law
in~\eqref{eq:bearing_fc}. This is done in a specific method as follows
\begin{lstlisting}
u = np.zeros(3)
if not self.is_leader:
    for j, neigh_pose in neigh_data.items():
   		err_pos = self.current.position - neigh_pose.position 
        err_vel = self.current_pose.velocity - neigh_pose.velocity
        u += - P_i[j] @ (self.kp * err_pos + self.kv * err_vel)
\end{lstlisting}
The rest of the \packagename/ ecosystem will handle ROS~2 communication among
nodes so that the user solely has to implement the desired control law. The
second required class needs to track the acceleration profile generated by the
distributed control scheme. This is developed according to
Section~\ref{sec:cf-control}. For this example, we need a control law tracking
the acceleration profile generated by the guidance class. Then, we generate an
angular velocity input according to a geometric control law. To this end, we
have to specify that the controller sends the Crazyflie angular rates and thrust
as control inputs.
This can be implemented as follows
\begin{lstlisting}
position_ctrl = FlatnessAccelerationCtrl()
attitude_ctrl = GeometryAttitudeCtrl()
sender = FPQRSender()
desired_accel = AccelerationTraj()

controller = HierarchicalController(
					pos_strategy=position_ctrl, 
					attitude_strategy=attitude_ctrl, 
					command_sender=sender, 
					traj_handler=desired_accel)	
rclpy.spin(controller)
\end{lstlisting}
The last class
instead simulates the Crazyflie dynamics.
In Figure~\ref{fig:class_comm}, we provide a \emph{communication diagram}
 highlighting the classes involved in this use-case and their interactions. For
 the formation control problem, 
  the guidance node of each quadrotor sends to its neighbors only two vectors
  (i.e., position and velocity) at $100$Hz rate.

\subsection{Running and Visualize the Simulation}
To easily run the simulation, we leverage the ROS~2 launching system.
Specifically, it is necessary to write a Python script specifying which nodes
have to be run. It is necessary to set-up instances of the
\texttt{DistributedControlGuidance}, \texttt{HierarchicalController} and
\texttt{CFIntegrator} classes for each quadrotor in the network. The launch file
also allows the user to specify additional parameters, e.g., robot initial
conditions and the communication graph. As an example, let $N$ be the number of
Crazyflie in the desired simulation. To instantiate a controller for each
quadrotor, the code can be formatted as follows.

\begin{lstlisting}
def generate_launch_description():
  # ... define graph and init. positions
  
  launch_description = [] # list overall nodes
  for i in range(N): # for each quadrotor add needed nodes
    # ... set-up distributed feedback node
    # ... set-up integrator node
    
    launch_description.append(Node(
		package='crazychoir_examples',
		node_executable='crazychoir_controller',
        namespace='agent_{}'.format(i),
        parameters=[{'cf_id': i}]))

  return LaunchDescription(launch_description)
\end{lstlisting}

\begin{figure}[h!]
\centering
	\includegraphics[width=.9\columnwidth]{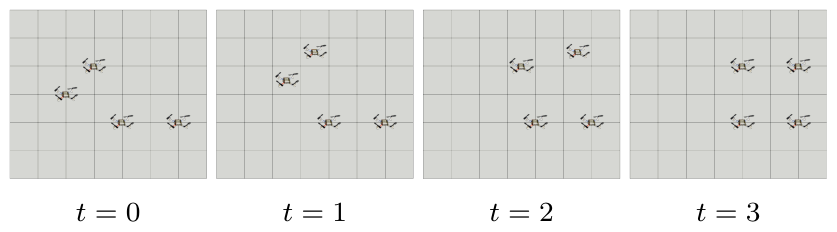}
	\caption{Sequence of images from the \textsc{Rviz} toolbox at different
	subsequent time instants. Static quadrotors are the leaders, while the moving
	ones are the followers. As time progresses, quadrotors reach the desired
	formation.}
	\label{fig:bearingfc}
\end{figure}

\begin{figure}[ht]
  \centering
	\includegraphics[height=3.8cm]{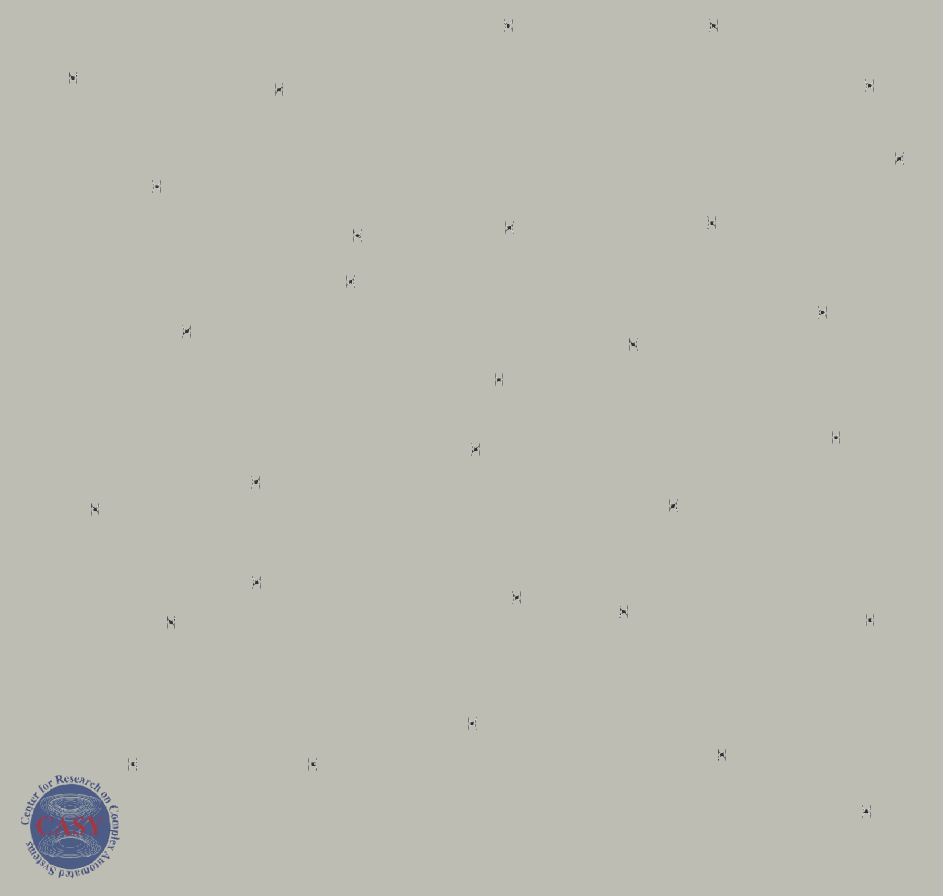}\hfill
	\includegraphics[height=3.8cm]{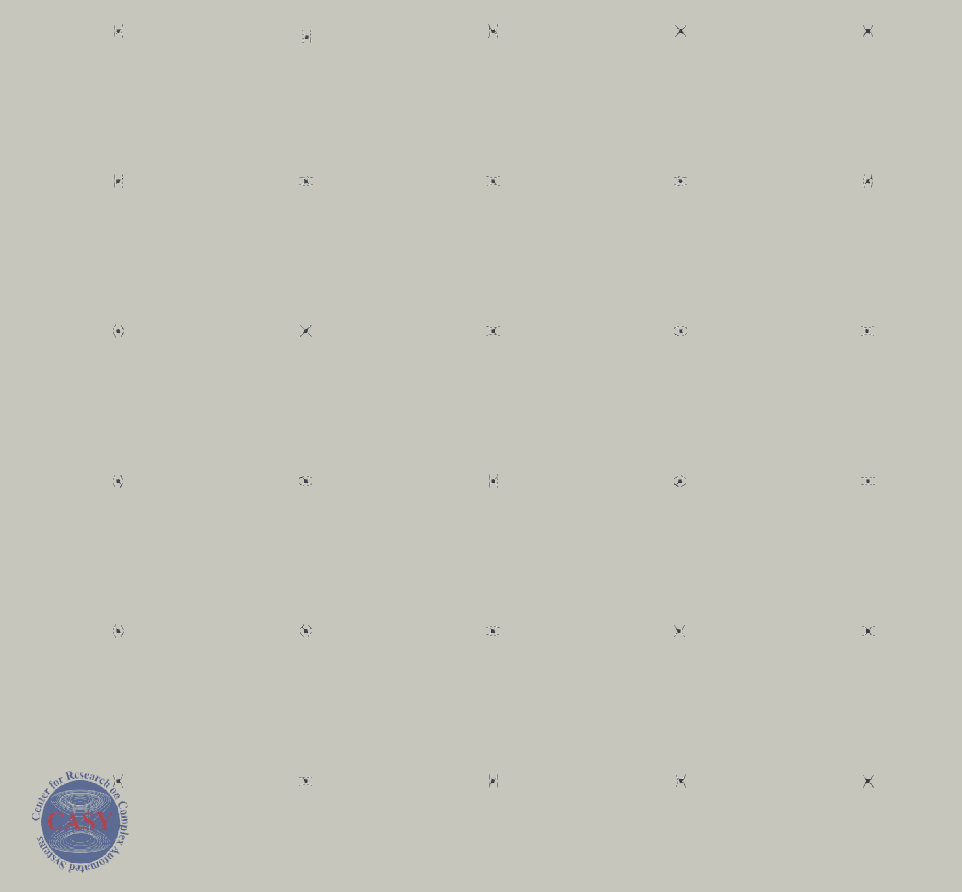}
	\caption{Beginning of the simulation (left) and end of the simulation (right)
	of the bearing formation control problem in Webots.}
	\label{fig:webots}
\end{figure}

In order to visualize the simulation, we included in our toolbox a script to
visualize the Crazyflie team in \textsc{Rviz}. More in detail, we provide a
class that receives the pose information from the \texttt{CFIntegrator} classes
and sends suitable messages to \textsc{Rviz} for visualization.
A set of snapshots for a simulation with $N=4$ robots is provided in
Figure~\ref{fig:bearingfc}.
To run the same setting in the Webots simulator, it is solely necessary to
substitute the \texttt{CFIntegrator} classes with the Webots simulation
architecture.
To emphasize the potential of Webots, we simulate $30$ quadrotors drawing a
  grid. In this setting, we leverage the distributed paradigm allowed by
  \packagename/ employing
  three workstations to handle the swarm, each of them endowed with Ubuntu 20.04
    and ROS~2 Foxy. One handles the physics engine, while each one handles a
    subset of the quadrotors. We implemented a keep-out zone to ensure that, as
    soon as a quadrotor flies beyond a certain limit area, it is immediately
    shut down. Snapshots from the Webots simulation are depicted in
    Figure~\ref{fig:webots}.

\subsection{Running the Experiment}
\begin{figure}[ht]
  \centering
	\includegraphics[width=.9\columnwidth]{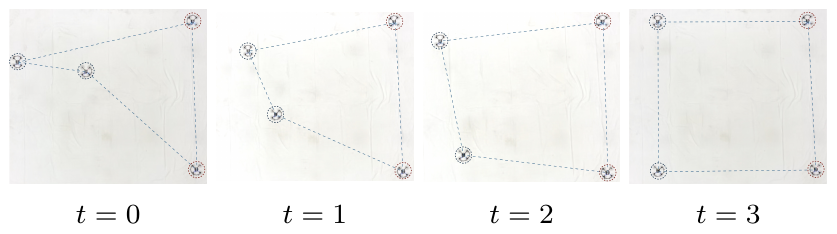}%
	\caption{Snapshots from an experiment at different time instants. Leaders are
  circled in red, followers in blue.}
	\label{fig:fc_experiment}
\end{figure}

In this section, we show how \packagename/ meets prototyping standards, allowing
the user to switch from simulation and experiment by changing only a few lines
of code. Indeed, to run an experiment, it is necessary to include in the launch
file the Crazyflie URIs and the radio nodes as described in
Section~\ref{sec:radio}. These few lines of code shall substitute the ones
related to Webots or \textsc{Rviz}. After that, it is only required to check
that nodes subscribe to the topics transporting Vicon data. 
To run our experiments, we used the \texttt{ros2-vicon-receiver} package
\footnote{\url{https://github.com/OPT4SMART/ros2-vicon-receiver}} to receive
data from our Vicon system. 
Snapshots from an experiment with $4$ quadrotor are provided in
Figure~\ref{fig:fc_experiment}.
Furthermore, we provide an experiment in which leaders track a hand-written
trajectory, leveraging the tools provided by the GUI (see
Section~\ref{subsec:gui}). More precisely, we draw a circle with a radius of
$0.4m$, and we execute this trajectory for  $40s$. To realize this scenario, we
changed only a few settings in the leader control node. We modified the control
strategy to \texttt{FlatnessPositionCtrl()}, which also processes position and
velocity references, and the trajectory handler to \texttt{FullStateTraj()}. In
addition, it is necessary to launch a planner node for the leaders, with the
goal of evaluating the splines retrieved by the drawing done in the GUI.
A video is available as supplementary material\footnote{The video is also
available at \url{https://youtu.be/mJ1HOquR-vE}}.

\section{\packagename/ Experiments on Trajectory Tracking and Pickup and Delivery}
\label{sec:experiment}
We now provide two additional experiments.
First, a quadrotor tracks a hand-written trajectory. Then, a swarm of Crazyflies
executes a pickup-and-delivery task. All the needed nodes are executed on a
workstation (Intel i9, Nvidia RTX 4000, Ubuntu 20.04, ROS~2 Foxy) equipped with
multiple radio transmitters, each handling two quadrotors. Moreover, a second
workstation (Intel Xeon, Windows 7) handles the Vicon Tracker software.

\subsection{Tracking of a hand-written trajectory}
\label{subsec:exp_hand-written}
\begin{figure}[h]
  \centering
  \includegraphics[width=.9\columnwidth]{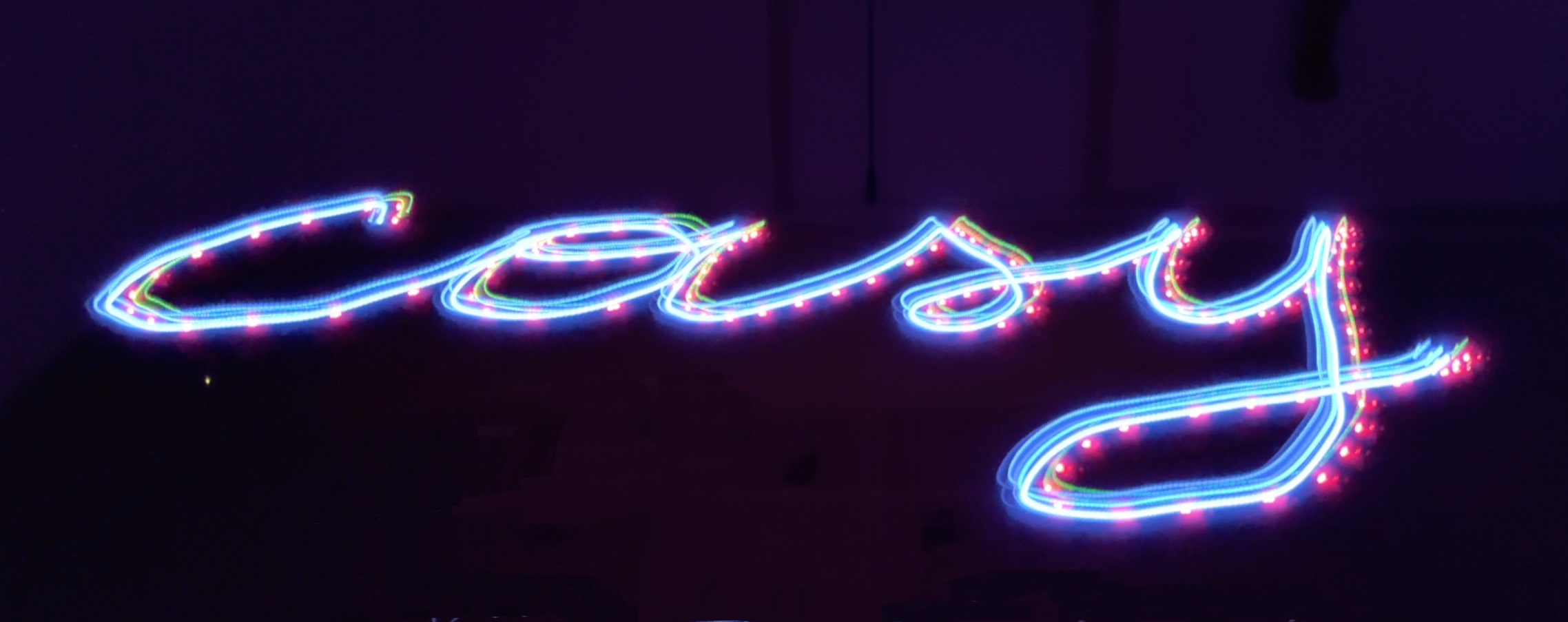}
	\caption{Long-term exposure snapshot from an experiment in which a quadrotor draws a hand-drawn trajectory.}
	\label{fig:single_experiment}
\end{figure}

We provide experimental results for a Crazyflie tracking a hand-drawn
trajectory. The path has been implemented using he graphical interface
introduced in Section~\ref{subsec:gui}, see also Figure~\ref{fig:gui}. 
The chosen trajectory, i.e., the name of our laboratory ``\textit{Casy}'', is
interpolated by a set of Python methods provided by the proposed planning
module. 
This trajectory is tracked by an off-board controller that elaborates the
desired references and, leveraging the \texttt{RadioFPQR} class, forwards the
thrust and angular-rates commands to the Crazyflie.
Figure~\ref{fig:single_experiment} shows a long-exposure snapshot from the
experiment. A video is available as supplementary material to the
paper\footnote{\label{note1}The video is also available at
\url{https://youtu.be/wX2r55WdZ9g}}.

\subsection{Multi-Robot Pickup and Delivery}
\label{sec:pdvrp}

By exploiting the distributed optimization features detailed in
Section~\ref{subsec:distr-opt}, we choose an experimental set-up that deals with
a distributed optimization problem.
\subsubsection{Problem Set-up}
We employ \packagename/ to perform experiments for a cooperative
pickup-and-delivery scenario involving $N=6$ Crazyflie. In this setting,
quadrotors have to pickup goods at given locations and delivery them in other
locations. As a further constraint, each quadrotor has a certain load capability
and can pick-up only some goods. The goal is to minimize the overall travel
distance, while guaranteeing that all the goods are successfully delivered. This
problem can be formulated as a large-scale mixed-integer linear program, i.e. an
optimization problem with a large number of both binary and continuous
variables. We refer the reader to~\cite{camisa2022multi} for a description of
the problem.

\subsubsection{Software Implementation}
\begin{figure}[h]
  \centering
  \includegraphics[width=0.48\columnwidth]{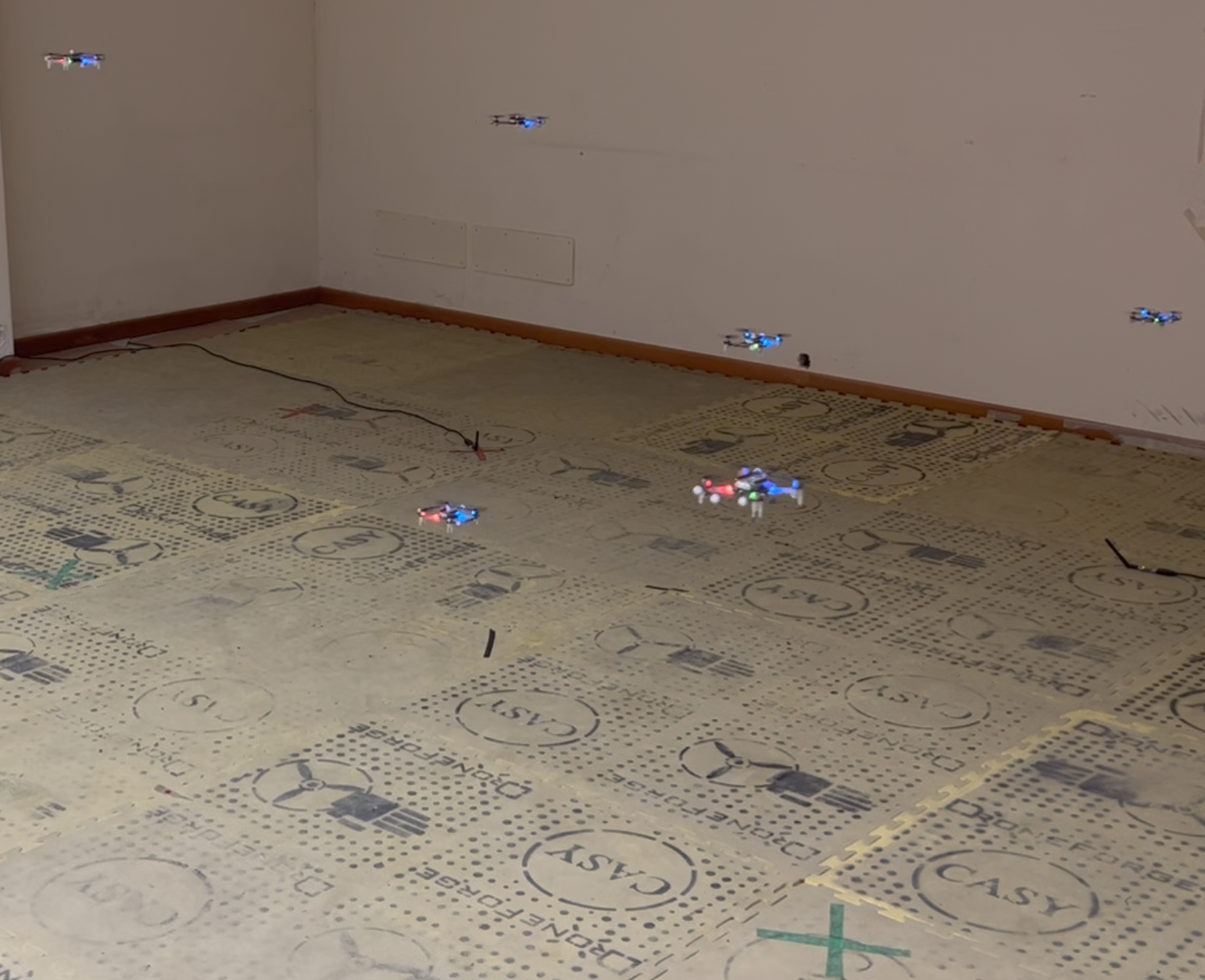}
  \includegraphics[width=0.48\columnwidth]{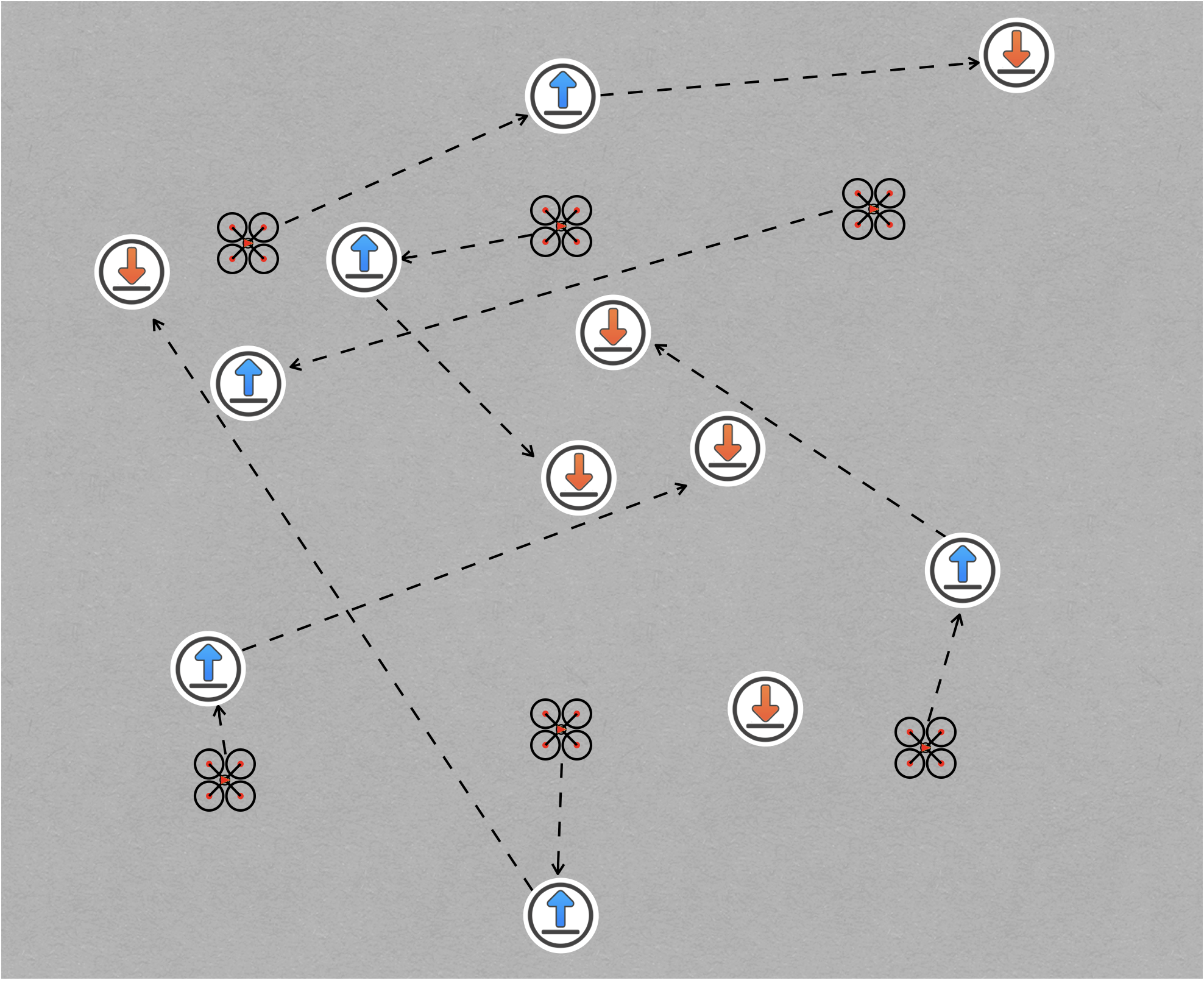}
	\caption{ The left snapshot illustrates the Crazyflies reaching the delivery
    locations, while the right scheme shows the pickup points (blue arrows) and
    the delivery points (orange arrows). Dashed lines represent optimal paths. }

	\label{fig:experiment}
\end{figure}

Similarly to Section~\ref{sec:simulation}, we exploit the modularity features of
\packagename/ to implement the considered scenario on a real swarm of
Crazyflies. Firstly, we take advantage of the \texttt{CFGuidance} class,
creating a child subclass named \texttt{PDVRPGuidance} which computes the
optimal pickup-and-delivery locations for each quadrotor. In this subclass, we
exploit the functionalities of \textsc{DISROPT}~\cite{farina2020disropt} to
model and solve the distributed optimization problem. 
As opposed to the simulative example provided in Section~\ref{sec:simulation},
in this experiment, we exploit the onboard controller, by sending position
references to each Crazyflie. Thanks to the \texttt{RadioHandlerXYZ} class, the
position references given by the solution of the pickup-and-delivery problem are
forwarded to the onboard controller of the Crazyflie, which is in charge to
generate and track the desired trajectory.
A snapshot from an experiment is in Figure~\ref{fig:experiment}. A video is
available as supplementary material\footnoteref{note1}.

\section{Conclusions}
\label{sec:Conclusions}
In this paper, we introduced \packagename/, a ROS~2 toolbox specifically
tailored for swarms of Crazyflie nano-quadrotors. The package allows the user to
perform realistic simulations of Crazyflie swarms using Webots and firmware
bindings. Also, the toolbox interfaces with radio dongles to perform experiments
on real nano-quadrotors. We provide several template classes to perform control,
planning, and cooperative decision making. Thanks to these functionalities,
users can easily extend the package by implementing their own algorithms.
Illustrative simulations and experiments have been provided to assess the
potentiality of the toolbox.

\bibliographystyle{IEEEtran}
\bibliography{biblio_ros2crazy}

\begin{thebibliography}{10}
\providecommand{\url}[1]{#1}
\csname url@samestyle\endcsname
\providecommand{\newblock}{\relax}
\providecommand{\bibinfo}[2]{#2}
\providecommand{\BIBentrySTDinterwordspacing}{\spaceskip=0pt\relax}
\providecommand{\BIBentryALTinterwordstretchfactor}{4}
\providecommand{\BIBentryALTinterwordspacing}{\spaceskip=\fontdimen2\font plus
\BIBentryALTinterwordstretchfactor\fontdimen3\font minus
  \fontdimen4\font\relax}
\providecommand{\BIBforeignlanguage}[2]{{%
\expandafter\ifx\csname l@#1\endcsname\relax
\typeout{** WARNING: IEEEtran.bst: No hyphenation pattern has been}%
\typeout{** loaded for the language `#1'. Using the pattern for}%
\typeout{** the default language instead.}%
\else
\language=\csname l@#1\endcsname
\fi
#2}}
\providecommand{\BIBdecl}{\relax}
\BIBdecl

\bibitem{quigley2009ros}
M.~Quigley, K.~Conley, B.~Gerkey, J.~Faust, T.~Foote, J.~Leibs, R.~Wheeler,
  A.~Y. Ng \emph{et~al.}, ``Ros: an open-source robot operating system,'' in
  \emph{ICRA workshop on open source software}, vol.~3, no. 3.2, 2009, p.~5.

\bibitem{maruyama2016exploring}
Y.~Maruyama, S.~Kato, and T.~Azumi, ``Exploring the performance of ros2,'' in
  \emph{13th Intern. Conf. on Embedded Software}, 2016, pp. 1--10.

\bibitem{preiss2017crazyswarm}
J.~A. Preiss, W.~Honig, G.~S. Sukhatme, and N.~Ayanian, ``Crazyswarm: A large
  nano-quadcopter swarm,'' in \emph{IEEE International Conference on Robotics
  and Automation (ICRA)}, 2017, pp. 3299--3304.

\bibitem{honig2017flying}
W.~H{\"o}nig and N.~Ayanian, ``Flying multiple uavs using ros,'' in \emph{Robot
  Operating System (ROS)}.\hskip 1em plus 0.5em minus 0.4em\relax Springer,
  2017, pp. 83--118.

\bibitem{shah2018airsim}
S.~Shah, D.~Dey, C.~Lovett, and A.~Kapoor, ``Airsim: High-fidelity visual and
  physical simulation for autonomous vehicles,'' in \emph{Field and service
  robotics}.\hskip 1em plus 0.5em minus 0.4em\relax Springer, 2018, pp.
  621--635.

\bibitem{song2020flightmare}
Y.~Song, S.~Naji, E.~Kaufmann, A.~Loquercio, and D.~Scaramuzza, ``Flightmare: A
  flexible quadrotor simulator,'' \emph{arXiv:2009.00563}, 2020.

\bibitem{grabe2013telekyb}
V.~Grabe, M.~Riedel, H.~H. B{\"u}lthoff, P.~R. Giordano, and A.~Franchi, ``The
  telekyb framework for a modular and extendible ros-based quadrotor control,''
  in \emph{IEEE ECMR}, 2013, pp. 19--25.

\bibitem{meyer2012comprehensive}
J.~Meyer, A.~Sendobry, S.~Kohlbrecher, U.~Klingauf, and O.~v. Stryk,
  ``Comprehensive simulation of quadrotor uavs using ros and gazebo,'' in
  \emph{IEEE SIMPAR}.\hskip 1em plus 0.5em minus 0.4em\relax Springer, 2012,
  pp. 400--411.

\bibitem{kumar2020erle}
K.~Kumar, S.~I. Azid, A.~Fagiolini, and M.~Cirrincione, ``Erle-copter
  simulation using ros and gazebo,'' in \emph{IEEE Mediterranean
  Electrotechnical Conference}, 2020, pp. 259--263.

\bibitem{silano2018crazys}
G.~Silano, E.~Aucone, and L.~Iannelli, ``Crazys: a software-in-the-loop
  platform for the crazyflie 2.0 nano-quadcopter,'' in \emph{IEEE Mediterranean
  Conference on Control and Automation}, 2018, pp. 1--6.

\bibitem{furrer2016rotors}
F.~Furrer, M.~Burri, M.~Achtelik, and R.~Siegwart, ``Rotors—a modular gazebo
  mav simulator framework,'' in \emph{Robot operating system (ROS)}.\hskip 1em
  plus 0.5em minus 0.4em\relax Springer, 2016, pp. 595--625.

\bibitem{nithya2019gazebo}
M.~Nithya and M.~Rashmi, ``Gazebo-ros-simulink framework for hover control and
  trajectory tracking of crazyflie 2.0,'' in \emph{IEEE Region 10 Conference},
  2019, pp. 649--653.

\bibitem{macenski2022robot}
S.~Macenski, T.~Foote, B.~Gerkey, C.~Lalancette, and W.~Woodall, ``Robot
  operating system 2: Design, architecture, and uses in the wild,''
  \emph{Science Robotics}, vol.~7, no.~66, p. eabm6074, 2022.

\bibitem{puck2020distributed}
L.~Puck, P.~Keller, T.~Schnell, C.~Plasberg, A.~Tanev, G.~Heppner, A.~Roennau,
  and R.~Dillmann, ``Distributed and synchronized setup towards real-time
  robotic control using ros2 on linux,'' in \emph{IEEE Intern. Conf. on
  Automation Science and Engineering}, 2020, pp. 1287--1293.

\bibitem{belsare2023micro}
K.~Belsare, A.~C. Rodriguez, P.~G. S{\'a}nchez, J.~Hierro, T.~Ko{\l}con,
  R.~Lange, I.~L{\"u}tkebohle, A.~Malki, J.~M. Losa, F.~Melendez \emph{et~al.},
  ``Micro-ros,'' in \emph{Robot Operating System (ROS) The Complete Reference
  (Volume 7)}.\hskip 1em plus 0.5em minus 0.4em\relax Springer, 2023, pp.
  3--55.

\bibitem{erHos2019ros2}
E.~Er{\H{o}}s, M.~Dahl, K.~Bengtsson, A.~Hanna, and P.~Falkman, ``A {ROS}2
  based communication architecture for control in collaborative and intelligent
  automation systems,'' \emph{Procedia Manufacturing}, vol.~38, pp. 349--357,
  2019.

\bibitem{reke2020self}
M.~Reke, D.~Peter, J.~Schulte-Tigges, S.~Schiffer, A.~Ferrein, T.~Walter, and
  D.~Matheis, ``A self-driving car architecture in {ROS}2,'' in \emph{IEEE
  International SAUPEC/RobMech/PRASA Conference}, 2020, pp. 1--6.

\bibitem{kaiserros2swarm}
T.~K. Kaiser, M.~J. Begemann, T.~Plattenteich, L.~Schilling, G.~Schildbach, and
  H.~Hamann, ``Ros2swarm-a ros 2 package for swarm robot behaviors,'' in
  \emph{IEEE ICRA}, 2022, pp. 6875--6881.

\bibitem{mai2022driving}
S.~Mai, N.~Traichel, and S.~Mostaghim, ``Driving swarm: A swarm robotics
  framework for intelligent navigation in a self-organized world,'' in
  \emph{IEEE Intern. Conf. on Robot. and Autom.}, 2022, pp. 01--07.

\bibitem{crazyswarm2}
``{Crazyswarm2},'' \url{https://github.com/IMRCLab/crazyswarm2/}.

\bibitem{testa2021choirbot}
A.~Testa, A.~Camisa, and G.~Notarstefano, ``Choi{R}bot: A {ROS} 2 toolbox for
  cooperative robotics,'' \emph{IEEE Robotics and Automation Letters}, vol.~6,
  no.~2, pp. 2714--2720, 2021.

\bibitem{michel2004cyberbotics}
O.~Michel, ``Cyberbotics ltd. webots™: professional mobile robot
  simulation,'' \emph{Int. Jour. of Adv. Robotic Sys.}, vol.~1, no.~1, p.~5,
  2004.

\bibitem{farina2020disropt}
F.~Farina, A.~Camisa, A.~Testa, I.~Notarnicola, and G.~Notarstefano, ``Disropt:
  a python framework for distributed optimization,'' \emph{IFAC-PapersOnLine},
  vol.~53, no.~2, pp. 2666--2671, 2020.

\bibitem{gamma1995design}
E.~Gamma, R.~Helm, R.~Johnson, R.~E. Johnson, J.~Vlissides \emph{et~al.},
  \emph{Design patterns: elements of reusable object-oriented software}.\hskip
  1em plus 0.5em minus 0.4em\relax Pearson Deutschland GmbH, 1995.

\bibitem{mellinger2011minimum}
D.~Mellinger and V.~Kumar, ``Minimum snap trajectory generation and control for
  quadrotors,'' in \emph{IEEE ICRA}, 2011, pp. 2520--2525.

\bibitem{lee2010geometric}
T.~Lee, M.~Leok, and N.~H. McClamroch, ``Geometric tracking control of a
  quadrotor uav on se (3),'' in \emph{IEEE conference on decision and control},
  2010, pp. 5420--5425.

\bibitem{kai2017nonlinear}
J.-M. Kai, G.~Allibert, M.-D. Hua, and T.~Hamel, ``Nonlinear feedback control
  of quadrotors exploiting first-order drag effects,''
  \emph{IFAC-PapersOnLine}, vol.~50, no.~1, pp. 8189--8195, 2017.

\bibitem{forster2015system}
J.~F{\"o}rster, ``System identification of the crazyflie 2.0 nano
  quadrocopter,'' {B.S.} thesis, ETH Zurich, 2015.

\bibitem{zhao2017translational}
S.~Zhao and D.~Zelazo, ``Translational and scaling formation maneuver control
  via a bearing-based approach,'' \emph{IEEE Transactions on Control of Network
  Systems}, vol.~4, no.~3, pp. 429--438, 2017.

\bibitem{camisa2022multi}
A.~Camisa, A.~Testa, and G.~Notarstefano, ``Multi-robot pickup and delivery via
  distributed resource allocation,'' \emph{IEEE Transactions on Robotics},
  vol.~39, no.~2, pp. 1106--1118, 2022.

\end{thebibliography}

\end{document}